\pdfoutput=1

\documentclass[11pt]{article}

\usepackage[]{acl}

\usepackage{times}
\usepackage{latexsym}

\usepackage[T1]{fontenc}

\usepackage[utf8]{inputenc}
\newcommand{\qafacteval}{\textsc{QAFactEval}}
\newcommand{\qafactevalnli}{\textsc{QAFactEval-NLI}}

\usepackage{graphicx}
\usepackage{amsmath}
\usepackage{breqn}
\usepackage{makecell}
\usepackage{tabularx}
\usepackage{booktabs}
\usepackage{multirow}
\usepackage[colorinlistoftodos]{todonotes}
\usepackage{textcomp}
\usepackage{bm}

\usepackage{hyperref}
\usepackage{amsmath}
\usepackage{multirow}
 
\usepackage{microtype}

\newcommand*{\affaddr}[1]{#1}

\title{QAFactEval: Improved QA-Based \\ Factual Consistency Evaluation for Summarization}

\author{
  \quad \textbf{Alexander R. Fabbri}
  \quad \textbf{Chien-Sheng Wu} \\
   \quad \textbf{Wenhao Liu}
  \quad \textbf{Caiming Xiong} \\
  \affaddr{Salesforce AI Research} \\
  \{afabbri, wu.jason, wenhao.liu, cxiong\}@salesforce.com
}

\begin{document}
\maketitle
\begin{abstract}
Factual consistency is an essential quality of text summarization models in practical settings.
Existing work in evaluating this dimension can be broadly categorized into two lines of research, entailment-based and question answering (QA)-based metrics, and different experimental setups often lead to contrasting conclusions as to which paradigm performs the best.
In this work, we conduct an extensive comparison of entailment and QA-based metrics, demonstrating that carefully choosing the components of a QA-based metric, especially question generation and answerability classification, is critical to performance. 
Building on those insights, we propose an optimized metric, which we call {\qafacteval}, that leads to a 14\% average improvement over previous QA-based metrics on the SummaC factual consistency benchmark, and also outperforms the best-performing entailment-based metric.
Moreover, we find that QA-based and entailment-based metrics can offer complementary signals and be combined into a single metric for a further performance boost.
\end{abstract}

\begin{table}[t!]
\centering
\small
\begin{tabularx}{\columnwidth}{|X X|}
\hline
\multicolumn{2}{|c|}{\textbf{Document}} \\
\multicolumn{2}{|>{\hsize=\dimexpr2\hsize+2\tabcolsep+\arrayrulewidth\relax}c|}{The Knicks beat \textcolor{blue}{the Rockets}. The fans were excited.}  \\ \hline
\multicolumn{2}{|c|}{\textbf{Summary}} \\
\multicolumn{2}{|>{\hsize=\dimexpr2\hsize+2\tabcolsep+\arrayrulewidth\relax}c|}{The Knicks beat \textcolor{red}{the Bucks}.}  \\ \hline
\multicolumn{1}{|c}{\multirow{1}{*}{\textbf{Entailment Matrix}}} & \multicolumn{1}{|c|}{\textbf{Selected Answer}} \\  
\multicolumn{1}{|c}{[Contra, Neutral, Support]} & \multicolumn{1}{|c|}{the Bucks} \\
\cline{2-2}
\multirow{4}{*}{$\left[ \begin{array}{ccc} 0.90 & 0.07 & 0.03 \\ 0.02 &  0.90 &  0.08 \end{array}\right]$} & \multicolumn{1}{|c|}{\textbf{Generated Question}}   \\  
  &  \multicolumn{1}{|X|}{Who did the Knicks beat?} \\  \cline{2-2}
  &  \multicolumn{1}{|c|}{\textbf{QA Output}} \\
  &  \multicolumn{1}{|c|}{the Rockets} \\    \hline
\multicolumn{1}{|c|}{\textbf{Max Support Score}} & \multicolumn{1}{c|}{\textbf{Answer Overlap Score}} \\
\multicolumn{1}{|c|}{0.08 }  &  \multicolumn{1}{c|}{0.20 } \\ \hline
\end{tabularx}
\caption{Toy example of a factual inconsistency between a summary and a source document.
{\textit{Left:}} The entailment-based metric computes the level of contradiction, neutrality, and support between the summary and each source document sentence.
The final factual consistency metric is calculated as the maximum support score over all source sentences. 
{\textit{Right:}} The QA-based metric first selects a noun-phrase \textit{answer} from the summary.
A QG model then generates an associated question that a QA model answers based on the source document. 
The answer overlap score of the QA-based metric measures the semantic overlap between the QA model output and the selected answer as the final metric score. 
}
\label{tab:example_intro}
\end{table}

\section{Introduction}\label{sec:introduction}
Text summarization aims to compress long document(s) into a short and fluent form that preserves salient information. 
The field has benefited from the application of pretrained methods~\cite{liu-lapata-2019-text,lewis2019bart,zhang2019pegasus}.
%
%
However, state-of-the-art models are not always factually consistent with the source documents they are conditioned on~\cite{maynez-etal-2020-faithfulness,fabbri2021summeval}. 
Thus, determining the factual consistency of a summary remains an essential task.
\par 
Recent metrics for summarization factual consistency can be broadly split into two categories: 
1) Entailment-based metrics that determine whether the content in the summary is entailed by the input document \cite{kryscinskiFactCC2019,DBLP:journals/corr/abs-2011-13662} and 
2) QA-based metrics that compute a factual consistency score based on a QA model's ability to answer, using the input document, questions generated from the summary \cite{DBLP:journals/corr/abs-2004-04228,durmus-etal-2020-feqa}. 
We provide an illustrative example in Table \ref{tab:example_intro} in which both metric types correctly identify the factual inconsistency and output a low score. 

Quantitative comparisons among entailment-based and QA-based metrics, however, often differ in their choices of baseline model and input granularity, evaluating on single datasets and drawing differing conclusions as to the best paradigm. 
For example, some work reports entailment-based metrics as performing best \cite{DBLP:journals/corr/abs-2011-13662,maynez-etal-2020-faithfulness}, while other work argues for QA metrics \cite{durmus-etal-2020-feqa,wang-etal-2020-asking,scialom2020QuestEval}.
Recently, \citet{laban2021summac} proposed a benchmark called \textit{SummaC} to compare metrics across six factual consistency datasets for the task of binary factual consistency classification, whether a summary is entirely factually consistent or not. 
This work unifies prior work on entailment-based metrics by studying the effect of input granularity, pretrained entailment model, and other hyperparameter choices on downstream evaluation performance. 
However, it does not study the components of QA-based metrics, which are more interpretable by their inherent decomposability.
\par
To unify work in QA-based factual consistency evaluation, we do an extensive hyperparameter analysis of current metrics. 
We break down these metrics into four constituent components: 1) the selection of answers to ask questions about, 2) question generation (QG) conditioned upon these answers, 3) question answering (QA) based on the source document, and 4) answer overlap evaluation between QA model output and selected answers. 
We study the effect of each of these components on metric performance. 
Based on our insights, we propose an optimized metric, which we call \qafacteval~, that outperforms the entailment-based metrics of \citet{laban2021summac}. 

Our contributions are the following: 
1) We analyze all components of the QA-based metric pipeline, and our proposed solution improves performance over prior QA-based metrics by over 14\% on a factual consistency benchmark consisting of 6 individual datasets, achieving state-of-the-art results. 
2) We show that QA-based metrics and NLI-based metrics offer complementary signals and combine them into a new metric via a simple learned network, further improving performance. 
3) We report results for 10 additional metrics across classification and correlation analysis, providing the most comprehensive benchmark results for factual consistency metrics and highlighting areas for future work in QA-based metrics~\footnote{Code and metric outputs will be made publicly available:
\url{https://github.com/salesforce/QAFactEval}}.
\section{Related Work}\label{sec:related-work}
\paragraph{Evaluating Factual Consistency}
Within entailment-based factual consistency evaluation,
\citet{falke-etal-2019-ranking} propose the task of ranking summary pairs for factual consistency based on entailment models, while
\citet{kryscinskiFactCC2019} explore factual consistency classification jointly with source support or contradiction span extraction. 
Other work on entailment-based metrics has examined input granularity \cite{goyal-durrett-2020-evaluating}, trained on adversarial datasets \cite{Barrantes2020AdversarialNF}, and explored entailment-based models as the backbone of others metrics such as BERTScore \cite{bert-score} as in \citet{koto2021ffci}. 
Metric comparisons, however, were often conducted on isolated datasets. 
\citet{laban2021summac} unify work in entailment-based metrics for factual consistency, showing the effect of granularity, base models, and other hyperparameter choices.
This work also proposes a learned metric built on top of the output of an entailment model, with parameters fine-tuned on synthetic data.
While this work fills a gap in the use of entailment-based metrics for factual consistency, our work analogously unifies QA-based metrics for factual consistency and proposes to combine entailment and QA-based metrics in a single learned metric.
\par
QA-based evaluation metrics have received attention for summary quality dimensions beyond factual consistency \cite{eyal-etal-2019-question,scialom-etal-2019-answers,deutsch2020questionanswering}.
Recent work has shown that QA-based metrics better measure the overlap of information units for determining summary relevance over embedding-based metrics \cite{deutsch-roth-2021-understanding}, further driving our study of QA-based metrics for factual consistency. 
%
%
While several QA-based metrics with similar structures have been applied for factual consistency,
\cite{durmus-etal-2020-feqa,wang-etal-2020-asking,scialom2020QuestEval}, they differ in their underlying answer selection, question generation, question answering, and answer overlap components, reporting different performances.
We perform a comprehensive evaluation of QA-based metric components and propose improved model components for the task of answer overlap and question filtering. 
\paragraph{Summarization Benchmarking}
A recent line of work aims to take stock of the current state of summarization models and progress, both within factual consistency and across summarization more broadly.
\citet{kryscinski-critical-eval} note biases and failure modes of abstractive summarization models, while other work analyzes and collects annotations over the output of recent summarization models across multiple dimensions, including factual consistency \cite{fabbri2021summeval,Bhandari-2020-reevaluating,huang-etal-2020-achieved}. 
\citet{lux-etal-2020-truth} propose a typology of errors found in summarization models, while \citet{gabriel-etal-2021-go} propose a framework for meta-evaluation of factual consistency metrics. \citet{laban2021summac} propose to combine recent work in factual consistency evaluation for summarization through a single benchmark. 
Our work directly makes use of this benchmark while emphasizing QA-based metrics. 
We also include correlation analysis for a more comprehensive understanding of current factual consistency metrics. 
\section{Evaluation Metrics}\label{sec:metrics}
In this section, we introduce the factual consistency metrics studied, which we divide into entailment metrics, QA-based metrics, and learned metrics. 
\subsection{Entailment-based Metrics}
We include the following entailment-based metrics due to further understand differences in granularity and base entailment models. 
The metrics below produce a score for each summary sentence that is then averaged to compute the final metric score. 
\paragraph{MNLI} applies a RoBERTa large \cite{liu2019roberta} model trained on MNLI \cite{williams-etal-2018-broad}. 
The score of a summary sentence is the maximum entailment score over all input sentences.
\paragraph{ANLI} \citet{Barrantes2020AdversarialNF} uses the same method as the MNLI metric with a model trained on the ANLI \cite{nie-etal-2020-adversarial} dataset consisting of adversarial datapoints.
\paragraph{SCZeroShot} \citet{laban2021summac} works analogously to the above metrics with a base model trained on both MNLI and Vitamin-C data \cite{schuster-etal-2021-get}, consisting of closely-related contrastive entailment examples. 

\paragraph{BertScore-FFCI} \citet{koto2021ffci} applies BertScore \cite{bert-score} with a backbone RoBERTa-MNLI model, averaging the three highest BertScore F1 scores over the input sentences. 
\paragraph{DAE} \citet{goyal-durrett-2020-evaluating} computes entailment scores between a source document and summary dependency arcs, applying an entailment model trained on synthetic data. 
\paragraph{FactCC} \citet{kryscinskiFactCC2019} is a RoBERTa-base model trained on FactCC synthetic data to compute a document-level score, and thus the scores need not be aggregated over input sentences.
\paragraph{DocNLI} \citet{yin-etal-2021-docnli} train a document-level entailment model, similar to the FactCC metric.
\subsection{QA Metric Components}
We now describe the components that constitute the QA-based pipeline for factual consistency. 
We refer to our metric, consisting of the best combination of the below components, as \qafacteval.
\paragraph{Answer Selection}
QA-based metrics compare information units between the summary and source, so it is thus necessary to first extract such units, or answers, from the given summary.
We follow the protocols from \citet{deutsch2020questionanswering} and compare extracting the following answer types: named entities (\textit{NER}), noun phrase chunks (\textit{NP Chunks}), maximally sized noun phrases (\textit{Max NP}), whereby the dependency subtrees of nouns reached by traversing a given sentence's dependency parse from the root are chosen as answers, and \textit{All}, which combines answers from the above three techniques. 
\paragraph{Question Generation}
Having selected answers, questions are generated conditioned upon these answers using the summary as context. 
Typically, this is an encoder-decoder model which inputs the answer and context separated by a special token. 
On the modeling side, we examine \textit{BART} \cite{lewis2019bart} and \textit{T5} \cite{2019t5} as the underlying generators. 
On the data side, we experiment with models trained for question generation on \textit{SQuAD} \cite{rajpurkar-etal-2016-squad}, a standard QA dataset consisting of questions on Wikipedia articles, and on \textit{QA2D} \cite{demszky2018transforming}, a dataset of declarative sentences with associated question/answer pairs derived from SQuAD. 
Furthermore, we experiment with the recently-introduced \textit{MixQG} models \cite{murakhovska2021mixqg}, which are T5 models trained on a combination of nine QA datasets with diverse answer types and which outperform other QG models across several tasks. 
We apply both the small and large versions of \textit{MixQG} to better understand the effect of QG model size. 
\paragraph{Question Answering}
The QA component answers questions from the previous steps using the input document as context. 
We experiment with both extractive QA models, which extract a text span from the input as an answer, and abstractive QA models, which generate an answer token-by-token.
For extractive models, we ablate \textit{Electra} \cite{clark2020electra}, a model architecturally similar to BERT \cite{devlin-etal-2019-bert} that achieves strong performance on the SQuAD 2.0 dataset and was previously used in measuring summary relevance \cite{deutsch2020questionanswering}. We also include \textit{MADE} \cite{friedman2021single}, which models multi-dataset QA with a collection of dataset-specific adapter modules sharing the same underlying RoBERTa-base model. 
For abstractive QA, we experiment with \textit{T5} fine-tuned on SQuAD and \textit{UnifiedQA} \cite{khashabi-etal-2020-unifiedqa}, an approach that trains a T5 QA model on 8 diverse, seed datasets and was shown to generalize across 20 datasets and 4 input formats. 
All QA models except MADE are trained on data containing unanswerable questions. 
Additional QA models can be included, although the above set of models allows us to inspect the aspects of interest in this study, namely extractive vs abstractive performance and multi-dataset training. 
\paragraph{Answer Overlap Evaluation}
An answer overlap metric must be computed to determine the match between the initial answer selected in the first component and the QA model output. 
Typically, answer overlap in QA is measured through exact match (\textit{EM}) score or word \textit{F1} score. 
We also test a learned metric, the \textit{LERC} score proposed by \citet{chen-etal-2020-mocha}.
This metric outputs a 1-5 answer overlap score conditioned on a question and context. The scorer is trained on their MOCHA dataset, consisting of 40k crowdsourced judgments on QA model outputs.
We include the BERT-base \cite{devlin-etal-2019-bert} model from the original paper, which we call \textit{LERC (orig)}. 
We additionally experiment with two models trained from RoBERTa-large checkpoints, one trained from the original checkpoint, \textit{LERC (RoBERTa)}, and one initialized from \citet{jia2021question}, which we call \textit{LERC (QuIP)}, for the task of jointly encoding passages and answers with question-infused pretraining. 
Lastly, we experiment with the \textit{IsAnsweredInput} answer metric, which is a 0/1 score of whether the question is answerable using the input document according to the QA model.
We use the Electra-large QA model to determine whether a question is answerable, as this model shows strong performance on identifying unanswerable questions on SQuAD.
%
%
%

%
\paragraph{Question Filtering}
Model-generated questions may contain noise from the QG model itself or from disfluencies in the summary the QG model conditions upon.
Such noisy questions can skew the overall metric score, as the QA component may be unable to correctly answer the question, regardless of the summary's factual consistency.
We filter such questions through a step called \textit{IsAnsweredSumm Filter}: the same Electra-large QA model returns a 0/1 score of whether the question is answerable, now using the summary as context, and questions labeled as unanswerable are filtered. 
\paragraph{Overall} 
For a given question, if IsAnsweredInput returns 0, the question is unanswerable using the input, we label all the above answer overlap scores as 0, and otherwise use the answer overlap score.
We refer to this scoring of unanswerable questions as 0 as the \textit{Answerability Penalty}.
We also experiment with not setting the overlap score of these unanswerable questions to 0 but rather using the answer overlap score of the most probable answer from the QA model.
Finally, the overall factual consistency score for each metric is computed as its average scores over all questions remaining following Question Filtering.
\subsection{Learned Metrics}
\paragraph{SCConv} is a model introduced by \citet{laban2021summac} that learns to aggregate entailment-model output scores across input sentences into a single score. 
More concretely, for a document consisting of $M$ sentences and a summary consisting of $N$ sentences, the entailment-based model produces an $M \times N$ matrix of entailment scores. 
The $M \times N$ matrix is then transformed to an $H \times N$ matrix by binning the $M$ sentences to create a histogram, where $H$ is the number of bins. 
This matrix is input to a 1-D convolution layer to produce a score for each summary sentence, and the scores are averaged across summary sentences. 
The parameters of this model are fine-tuned on synthetic data, detailed in Section \ref{sec:experiment_setup}
\paragraph{\qafactevalnli}
While SCConv captures sentence-level support, \qafacteval~ measures finer-grained answer overlap between the source and summary. 
Thus, we are able to combine these two into a single factual consistency metric, \qafactevalnli. 
Assume that $K$ answers are extracted from the summary. 
The pipeline described above will then output a single score per answer for the entire summary, resulting in an array of length $K$. 
We convert this to a histogram of size $H$ in a similar manner as SCConv and pass this histogram through a 1-D convolution layer to produce a single QA score.
This score is concatenated with the NLI score produced by SCConv and input to a linear layer to produce the final metric score. 
The linear layer can be trained in either \textit{synthetic} or \textit{supervised} ways, detailed in Section \ref{sec:experiment_setup}.

\subsection{Additional Metrics}
We include the following metrics for completeness. 
\paragraph{BARTScore} \citet{yuan2021bartscore} calculates the log-likelihood from BART fine-tuned on CNN/DailyMail \cite{hermann2015teaching,nallapati-etal-2016-abstractive} of the summary conditioned upon the source text as a metric for factual consistency. 
\paragraph{BLANC} \citet{vasilyev-etal-2020-fill} is a reference-less metric of summary quality that measures the difference in masked language modeling performance with and without access to the summary. 
\paragraph{QuestEval} \cite{scialom2020QuestEval} is the prior state-of-the-art QA-based metric for factual consistency. 
The T5-base (SQuAD) QG and T5-base QA models described above are applied directly from the QuestEval metric.
QuestEval generates questions based on the input document and answers them using the summary in addition to following the above QA metric pipeline. 
QuestEval aggregates the score from these two pipelines. 
We believe that our described pipeline more closely measures factual consistency, while generating questions from the source may confound factual consistency with relevance.

\section{Methodology}\label{sec:methodology}
We present the datasets explored for binary classification and correlation analyses. 
We also describe settings for reporting ablation and final results.
\subsection{Data}
The \textbf{SummaC} benchmark \cite{laban2021summac} introduces a collection of datasets for binary factual consistency evaluation. 
A data point is labeled as positive if it contains no factual inconsistencies or is rated the highest possible score in the case of Likert scaling, and as negative otherwise. 
We now briefly describe the datasets in the benchmark and any departures from the original benchmark, and additional datasets we use for correlation analysis. 
We refer the reader to \citet{laban2021summac} for further details regarding the benchmark creation. 
\paragraph{CGS} \citet{falke-etal-2019-ranking} consists of paired summary sentences from CNN/DailyMail \cite{hermann2015teaching,nallapati-etal-2016-abstractive}, one correct sentence and one containing an error.
\citet{laban2021summac} treats the correct summaries as positive examples and the others as negative examples.
\paragraph{XSF} \citet{maynez-etal-2020-faithfulness} consists of summaries from the XSum dataset \cite{Narayan2018DontGM} annotated for word-level factual consistency errors. 
\paragraph{Polytope} \citet{huang-etal-2020-achieved} propose a typology of eight summarization errors consisting of both content and stylistic errors and annotate model outputs from 10 systems on CNN/DailyMail data.
The original SummaC benchmark included the Omission and Addition errors of this proposed typology as factual inconsistencies, but these are largely extractive, factually consistent summaries.
We thus label these examples as factually consistent and report results on this modified dataset.
\paragraph{FactCC} \citet{kryscinskiFactCC2019} introduce a factual consistency dataset on CNN/DailyMail annotated by the authors of the paper to ensure the quality of the annotations. 
\paragraph{SummEval} \citet{fabbri2021summeval} analyze summaries from 17 models on CNN/DailyMail across the dimensions of factual consistency, coherence, fluency, and relevance. 
\paragraph{FRANK} \citet{pagnoni-etal-2021-understanding} introduce an extensive typology of errors made by summarization systems across CNN/DailyMail and XSum.
\paragraph{QAGs} \citet{wang-etal-2020-asking} crowdsource sentence-level summary annotations for factual consistency across CNN/Daily Mail and XSum data. 
We only report correlation analysis for this dataset as it was not a part of SummaC.
\subsection{Experiment Setup}
\label{sec:experiment_setup}
\paragraph{Metric Implementation}
Metrics were applied directly from the original GitHub repository or by using the SacreRouge Library \cite{deutsch-roth-2020-sacrerouge}, which was also used in correlation analysis. 
The learned metrics make use of code released from \citet{laban2021summac} for training, and all models are implemented in PyTorch ~\citep{paszke2020pytorch} and in the Transformers library \cite{wolf2019huggingface}.
The BART-large (QA2D) QG and Electra-large QA models are applied from the QAEval relevance modeling metric \cite{deutsch2020questionanswering}. 
\paragraph{Ablation Settings}
Following \citet{laban2021summac}, a metric threshold score for binary classification is determined from the validation set of SummaC and applied to the test set. 
This threshold score is determined for every metric studied. 
Furthermore, we note that hyperparameter choices for several of the strong entailment baselines, namely SCConv, SCZeroShot, and MNLI are derived from \citet{laban2021summac}, thus providing a reasonable comparison to QAFactEval, whose hyperparameters we tune on the SummaC validation set. 
For ablation studies, we both perform thresholding and evaluation on the validation set to preserve the integrity of the test set. 
For each benchmark dataset, we sample a random subset of 80\% of the validation set to determine the threshold and evaluate on the remaining 20\% of the validation set. 
The best performing combination of QA metric components constitutes our \qafacteval~ metric. 
We take the best performing combination of QA metric components and vary a given component, such as answer selection, while holding all other components constant and consistent with the best component combination. 
\paragraph{Training Settings}
To tune the parameters of the learned metrics, we train on a subset of 50k synthetic data points from FactCC, following \citet{laban2021summac}.
We name these runs \textit{synthetic} setting due to the lack of human-labeled data. 
We also experiment with a \textit{supervised} setting by fine-tuning the parameters on the SummaC validation set for each individual dataset, choosing the threshold on this validation data, and applying the model to the test set.
Training on such a small amount of data is feasible due to the small number of parameters of the learned metrics. 
Cross entropy loss with Adam~\cite{Adam} optimizer is used, with a batch size of 32 and a learning rate of 1e-2. 
\section{Results}\label{sec:results}
In this section, we first study the effects of model component choices on \qafacteval~.
We then compare metric results across both the SummaC binary classification task and correlation analysis.
\subsection{Ablation Results}

We provide the results of our ablation studies on the components of QA-based metrics in Table \ref{tab:ablation_all} and show two illustrative examples in Table \ref{tab:example_qg}. 
\paragraph{Effect of Answer Selection}
Selecting NP Chunks performs best, aligning with \citet{deutsch2020questionanswering}, which shows that NP Chunks obtain the largest coverage of information units while retaining high precision.
We find a large decrease in performance when selecting NER and only a slight decrease in performance when choosing Max NP or ALL answers together. 
Named entity selection likely performs worse due to the scarcity of extracted answers; only three entities are extracted on average across the benchmark, while all other approaches extract over 10 answers per summary. 
\paragraph{Effect of QG Models}
\begin{table}[t!]
\centering
\tiny
\resizebox{\columnwidth}{!}{\begin{tabular}{|l l c|}
\hline
    \textbf{Component}   & \textbf{Model Choice} &  \textbf{Benchmark}    \\ \hline
     \textbf{\qafacteval~}   &  & \textbf{77.5}   \\   \hline
    \multirow{4}{*}{Answer Selection}   &  \textbf{NP Chunks} &  -  \\  
      & Max NP &  75.7  \\  
      & NER     &  66.4 \\ 
      & ALL    & 75.7 \\ \hline
    \multirow{5}{*}{Question Generation}   &  \textbf{BART-large (QA2D)} & - \\ 
     & BART-large (SQuAD) &  74.3  \\  
      & T5-base (SQuAD)     &  67.0  \\ 
      & MixQG-base    & 75.1   \\ 
      & MixQG-large   & 74.9  \\ \hline
    \multirow{4}{*}{Question Answering}   &  \textbf{Electra-large} & - \\
      & Electra-base &  77.0 \\  
      & MADE &  77.4 \\  
      & \textit{T5-base}  &  76.1  \\ 
      & \textit{UnifiedQA-base}   & 75.7   \\ \hline
     \multirow{5}{*}{Answer Overlap}   & \textbf{LERC (QuIP)} &  -   \\ 
      & EM &  68.4   \\  
      & F1 &   71.7  \\ 
      & IsAnsweredInput &  73.3    \\ 
      & LERC (orig) &  71.8    \\ 
       & LERC (RoBERTa) &  77.3    \\ \hline
      \multirow{4}{*}{Filtering/Answerability}  &  \textbf{Both} & - \\
         & No IsAnsweredSumm Filter &  73.8   \\  
         & No Answerability Penalty &  72.1   \\  
         & Neither &  67.4   \\  \hline
\end{tabular}}
\caption{Results of ablation studies on the SummaC benchmark validation set, showing the effect of the individual components of \qafacteval~.
The first row represents the performance of the best combination of components. 
Ablations are performed by swapping a given component while holding all others consistent with the best overall model, and the best setting is bolded.}
\label{tab:ablation_all}
\end{table}

\begin{table*}[t!]
\resizebox{\textwidth}{!}{\begin{tabular}{|l|l|c|c|c|c|c|c||c|}
\hline
    \textbf{Model Type}   & \textbf{Model Name}  & \textbf{CGS} & \textbf{XSF} & \textbf{Polytope} &  \textbf{FactCC} & \textbf{SummEval} & \textbf{FRANK} &  \textbf{Benchmark}    \\ \hline
    \multirow{2}{*}{Misc}         & BARTScore     &  63.3 & 53.3 & 80.4 & 66.8 & 69.8 & 80.0 & 68.9  \\ 
                                  & BLANC     & 51.6 & 54.5 & 72.2 & 53.0 & 63.0 & 76.2 & 61.8  \\  \hline
                                  & FactCC     &   64.8 & 56.6 & 80.2 & 77.1 & 73.6 & 70.3 & 70.4  \\  
    \multirow{6}{*}{Entailment}   & BertScore-FFCI     & 56.9 & \textbf{68.8} & 69.2 & 57.9 & 67.4 & 71.9 & 65.4  \\ 
                                  & DAE     &  71.3 & 49.7 & 78.9 & 80.7 & 74.7 & 81.0 & 72.7   \\ 
                                  & ANLI     &  74.9 & 53.0 & 77.6 & 85.8 & 75.9 & 78.9 & 74.4    \\ 
                                  & MNLI     &  67.6 & 61.5 & 77.3 & 89.8 & 78.7 & 79.6 & 75.7    \\ 
                                  & DocNLI     &  49.6 & 57.0 & \textbf{84.7} & 73.0 & 75.6 & 70.9 & 68.5  \\ 
                                  & SCZeroShot     &  59.6 & 56.1 & 81.5 & 83.2 & 77.9 & 78.5 & 72.8 \\ \hline
    \multirow{2}{*}{QA}           & QuestEval  &  59.4 & 61.9 & 73.1 & 66.5 & 68.4 & 79.8 & 68.2   \\ 
                                  & \qafacteval~   &  75.1 & 63.1 & 79.8 & 84.1 & \textbf{80.9} & 83.9 & 77.8  \\ \hline
    \multirow{4}{*}{Learned}     & SCConv (synthetic)  & 60.8 & 60.9 & 76.0 & 88.1 & 78.1 & 81.6 & 74.3  \\ 
                                  & \qafactevalnli~(synthetic)  & 74.2 & 59.1 & 82.1 & \textbf{91.1} & 80.2 & 83.4 & 78.3  \\
                                 & \qafactevalnli~(supervised) & \textbf{78.1} & 60.9 & 83.7 & 89.3 & 80.5 & 
                                 \hspace{1mm} \textbf{84.3*} & \hspace{1mm} \textbf{79.5*}  \\  \hline
\end{tabular}}
\caption{Balanced accuracy on the test set of the six SummaC benchmark datasets, and the average over the benchmark. Metrics are divided into entailment-based, QA-based, and learned metrics that are fine-tuned on synthetic or supervised data. An improvement over prior work with a 99\% confidence interval is indicated by *.}
\label{tab:overall_results}
\end{table*}
%

\begin{table*}[t!]
\centering
\small
\begin{tabularx}{\textwidth}{X||X|X|X|X}
\hline
\multicolumn{1}{|c}{\textbf{Document}} & \multicolumn{2}{|>{\hsize=\dimexpr2\hsize+1\tabcolsep+\arrayrulewidth\relax}X|}{
Paul Merson has restarted his row with Andros Townsend. ... '... it was a great goal,' Merson said. 'It's just a matter of opinion, and ... he got pulled off after half an hour .... in front of Roy Hodgson, \textcolor{purple}{so he shouldn't have been in the squad.} ...' ... \textcolor{blue}{Sky Sports pundit  Merson (centre) criticised Townsend's call-up to the England squad last week .}... } & \multicolumn{2}{|>{\hsize=\dimexpr2\hsize+1\tabcolsep+\arrayrulewidth\relax}X|}{ They're not gonna take it anymore. Really. Twisted Sister says that its 2016 tour will be its last, according to a press release. ... \textcolor{blue}{The band will also perform two shows} in Pero's honor: one at Las Vegas Hard Rock Hotel and Casino, the other at the Starland Ballroom in Sayreville, New Jersey. } \\ \hline
\multicolumn{1}{|c}{\textbf{Summary}} & \multicolumn{2}{|>{\hsize=\dimexpr2\hsize+1\tabcolsep+\arrayrulewidth\relax}X|}{Paul Merson is not happy with Andros Townsend's call-up to the England squad last week}  & \multicolumn{2}{|>{\hsize=\dimexpr2\hsize+1\tabcolsep+\arrayrulewidth\relax}c|}{The band will perform two shows.} \\ \hline
%
\multicolumn{1}{|c}{\textbf{Selected Answer}} & \multicolumn{2}{|>{\hsize=\dimexpr2\hsize+1\tabcolsep+\arrayrulewidth\relax}c|}{Andros Townsend's call-up} & \multicolumn{2}{|>{\hsize=\dimexpr2\hsize+1\tabcolsep+\arrayrulewidth\relax}c|}{the band} \\ \hline
%
\multicolumn{1}{|c}{\textbf{Question Generation}} & \multicolumn{1}{|c}{\textbf{BART-QA2D}} & \multicolumn{1}{|c|}{\textbf{MixQG-large}} & \multicolumn{2}{>{\hsize=\dimexpr2\hsize+1\tabcolsep+\arrayrulewidth\relax}c|}{\textbf{BART-QA2D Question}} \\ 
\multicolumn{1}{|c|}{}  & What is Paul Merson not happy with to the England squad last week? & What is Paul Merson not happy with? & \multicolumn{2}{>{\hsize=\dimexpr2\hsize+1\tabcolsep+\arrayrulewidth\relax}c|}{\multirow{2}{*}{Who will perform two shows?}} \\ \hline
\multicolumn{1}{|c|}{ \textbf{QA Output}} & Townsend's call-up  & he shouldn't have been in the squad & \multicolumn{2}{|>{\hsize=\dimexpr2\hsize+1\tabcolsep+\arrayrulewidth\relax}c|}{Unanswerable (Twisted Sister)} \\ \hline
 \multicolumn{1}{|c|}{\textbf{Answer Overlap}} &  \multicolumn{1}{|c|}{$1.00$}  &  \multicolumn{1}{|c|}{$0.30$} & \multicolumn{2}{|>{\hsize=\dimexpr2\hsize+1\tabcolsep+\arrayrulewidth\relax}c|}{ $0.00$ ($0.80$)}  \\ \hline
\end{tabularx}
\caption{Example source documents and summaries along with QA-based metric component outputs. \textit{Left:} This example illustrates that the fluency of the QG model does not necessarily improve downstream factual consistency evaluation performance; the less fluent, more extractive BART-QA2D question is more-easily answerable by the QA model. Not shown, the entailment-based SCConv metric incorrectly labels this entity-centric example, likely due the introduction of novel unigrams.  \textit{Right:} The QA model incorrectly labels this question as unanswerable, perhaps due to the generality of the question or due to noise in the input document. The QA output and our learned overlap score if forced to extract an answer are in parenthesis. SCConv correctly labels this highly extractive example.}
\label{tab:example_qg}
\end{table*}

The choice of the QG model notably affects downstream performance. 
BART-large (QA2D) works the best and produces much longer questions, about 17 tokens on average, versus about 10 from the other models.
\citet{deutsch2020questionanswering} note how humans tend to produce shorter questions. However, longer questions may be preferable for this task to facilitate the QA model's ability to understand and answer the question. 
BART-large (QA2D) also is the most extractive, with only about 20\% novel unigrams in the question, while T5-base (SQuAD) model is the most abstractive with about 47\% novel unigrams, resulting in occasional hallucinations and questions that the QA model struggles to answer.  
\begin{table*}[t!]
\resizebox{\textwidth}{!}{\begin{tabular}{|c|c|c|c|c|c|c|c|}
\hline
    \textbf{Model Type}   & \textbf{Model Name} & \textbf{XSF} & \textbf{SummEval} & \textbf{FRANK-CNNDM} &  \textbf{FRANK-XSum} & \textbf{QAGs-CNNDM} &  \textbf{QAGs-XSum}    \\ \hline
    \multirow{2}{*}{Misc}         & BARTScore     & 0.25 & 0.37 & 0.58 & 0.15 & \textbf{0.73} & 0.17  \\
                                  & BLANC     & 0.03 & 0.20 & 0.33 & 0.07 & 0.33 & 0.02 \\ \hline
                                  & FactCC     &  0.04 & 0.37 & 0.38 & 0.06 & 0.40 & 0.30   \\ 
    \multirow{6}{*}{Entailment}   & BertScore-FFCI     & \textbf{0.45} & 0.27 & 0.36 & 0.16 & 0.53 & 0.21    \\ 
                                  & DAE     &  0.02 & 0.45 & 0.50 & 0.22 & 0.63 & -0.20    \\ 
                                  & ANLI     &  0.16 & 0.43 & 0.53 & 0.18 & 0.65 & 0.39  \\ 
                                  & MNLI     &  0.18 & 0.44 & 0.52 & 0.18 & 0.66 & 0.35   \\ 
                                  & DocNLI     &  0.01 & 0.41 & 0.12 & 0.26 & 0.16 & -0.34   \\ 
                                  & SCZeroShot     &  0.06 & 0.50 & 0.55 & 0.27 & 0.57 & \textbf{0.44} \\ \hline
    \multirow{2}{*}{QA}           & QuestEval  &  \textbf{0.45} & 0.41 & 0.52 & 0.24 & 0.51 & 0.23  \\ 
                                  & \qafacteval~    & 0.29 & \textbf{0.61} & \textbf{0.66} & \textbf{0.32} & \textbf{0.68} & 
                                  \textbf{0.44}   \\ \hline
    \multirow{2}{*}{Learned}     & SCConv (synthetic)  &  0.12 & 0.50 & 0.59 & \textbf{0.30} & 0.03 & 0.06    \\ 
                                  & \qafactevalnli (synthetic)    &  0.19 & \textbf{0.61} & \textbf{0.66} & 0.25 & 0.65 & \textbf{0.48}    \\ \hline
\end{tabular}}
\caption{Instance-level Pearson correlation coefficients across factual consistency evaluation datasets. Metrics are divided into entailment-based, QA-based, and learned metrics that are fine-tuned on synthetic or supervised data. The two highest-correlated metrics for each dataset are shown in bold.}
\label{tab:corr_glob_pearson}
\end{table*}
As seen in Table \ref{tab:example_qg}, MixQG models do often produce highly-fluent questions, but the longer, highly-extractive output of BART-large (QA2D) improves downstream factual consistency performance. 
\paragraph{Effect of QA Model}
Surprisingly, we do not find a large difference in the QA model component across model sizes or between extractive and abstractive QA models, implying that QA ability is not the bottleneck of our task. 
In this setting, we keep IsAnsweredInput from Electra-large constant, as not all QA models are trained with unanswerable questions; thus the only differences are in the answers to questions marked as answerable. 
\paragraph{Effect of Answer Overlap Metric}
We observe a large difference between EM and other overlap metrics. 
We also see a notable gap between LERC (orig) and LERC (RoBERTa) along with a further slight improvement with LERC (QuIP), showing the effect of the underlying model of the learned metric on factual consistency performance. 
\paragraph{Effect of Question Filtering and Answerability}
Not filtering questions according to the QA model's ability to answer them conditioned upon the summary decreases performance. 
Furthermore, not applying the Answerability Penalty, and using the answer overlap score for the most probable answers for all questions, even those judged unanswerable by the QA model, also decreases performance. 
While the answer overlap metric should capture unanswerable questions for information not found in the input (extrinsic error), the selected answer may appear in both the summary and source but in different contexts (intrinsic error).
The QA model may return this as the most probable answer and be scored as correct by the answer overlap component despite a factual inconsistency. 
This finding demonstrates the importance of determining question answerability, a point also emphasized in \citet{deutsch2020questionanswering} for QA-based metrics of relevance. 
Removing both of these components results in a drastic performance decrease.
\subsection{Overall Results}
We present the results on the test set of SummaC in Table~\ref{tab:overall_results}. 
\qafacteval~ shows a substantial improvement over the previous state-of-the-art QA metric for factual consistency, QuestEval. 
Furthermore, it outperforms all other entailment-based metrics. 
\qafactevalnli~ shows slight improvements on the \textit{synthetic} data. 
Notable improvements in this synthetic setting can be observed on the FactCC dataset, likely as the synthetic FactCC data the model is trained on was designed to mirror the errors captured in annotations. 
This performance boost on FactCC motivated our use of \textit{supervised} data for fine-tuning our learned metric. 
Supervised fine-tuning on validation data helps in most cases and \qafactevalnli~(supervised) improves on the overall benchmark by a statistically significant margin, using bootstrap resampling \cite{Efron} with Bonferroni correction \cite{bonferroni1935calcolo} to obtain 99\% confidence intervals (see Appendix for details).
The performance drop on FactCC could be due to the proximity of the synthetic data to the labeled data and the data size difference. 
BertScore-FFCI performs best on XSF perhaps due to the closeness between BertScore's token-level metric and XSF's word-level annotations, and DocNLI's Polytope performance may also be from training data similarity.

We find that {\qafacteval} and SCConv do offer complementary signals that can be learned from supervised data.
Individually fine-tuning the learned SCConv or a learned variation of {\qafacteval} on supervised data did not improve results over the non-supervised metrics; this result suggests the necessity of combining the two for further improvements.  
Training on the validation sets combined, rather than on each individual dataset separately, did not give an improvement, likely due to the learnable combination of NLI and {\qafacteval} being dataset dependent. 
\subsection{Correlation Analysis}
We provide instance-level Pearson correlation between aggregated human judgments and metric scores for each model to compare to previous work in factual consistency that reports correlation analysis. 
Results are shown in Table \ref{tab:corr_glob_pearson}.
We split FRANK into CNN/DailyMail and XSum subsets for finer-grained analysis, as substantial differences have been noted in correlation performance across the two datasets \cite{durmus-etal-2020-feqa}. 
We exclude Polytope, FactCC, and CGS here as prior work has only studied these datasets for binary classification. 

We find that \qafacteval~ performs well across most datasets.
As in the classification results, BertScore-FFCI's performs well on XSF, and we note that QuestEval's answerability classifier correlates more so with these fine-grained annotations than on other datasets.
\qafactevalnli~performs well on most datasets except XSF. 
Fine-tuning on FactCC synthetic data for binary classification likely does not capture the aggregated, word-level factuality scores of XSF. 
We leave a study of fine-tuning this model on supervised data with a regression loss for future work.
\section{Conclusion}\label{sec:conclusion}
In this work, we demonstrated that QA-based metrics, when its components are properly optimized, outperform entailment-based metrics on a comprehensive factual consistency evaluation benchmark. 
We identify question generation and answerability detection as key components for improving QA-based metrics in future work. 
Furthermore, we show that entailment and QA-based metrics offer complementary signals through a combined metric that achieves state-of-the-art performance on this benchmark. 
We believe that our work lays the foundation for future work in QA-based metrics for factual consistency by offering a fairer comparison to other metrics across datasets and settings. 
\section{Ethical Considerations}

\paragraph{Dataset Biases}
The underlying models of the metrics presented in this work are trained on documents in English and thus mainly represent the culture of the English-speaking populace. 
Political or gender biases may also exist in the datasets, and models, and subsequently the metrics, trained on these datasets may propagate these biases.
We did not stress test these metrics for such biases and request that the users of these metrics be aware of these potential issues in applying them. 
\paragraph{Misuse Potential and Failure Mode}
When properly used, the metrics described in this paper can be a useful tool for detecting summarization model errors.
However, the current metrics fail to detect all factual inconsistencies, which must be remembered when applying these metrics as a filter for downstream applications. 
Factual inconsistencies in summaries could contribute to misinformation on the internet. 
\paragraph{Environmental Cost}
The experiments described in the paper primarily make use of A100 GPUs. 
Most of the metrics have already been trained, in which case we simply ran inference using the existing models. 
We typically used a single GPU per experiment. 
Training learned answer overlap components can take a couple of hours, while experiments for learned metrics on SummaC take less than 10 minutes. 
These are the base models used in these experiments, with the number of parameters, in millions, in parentheses: BERT-base (110), BART-large (400), Electra-base (110), Electra-large (335), RoBERTa-large (355), T5-base (220), T5-large (770). 
Future work may analyze the effect of using distilled backbone models on factual consistency evaluation. 
\bibliography{anthology,custom}
\bibliographystyle{acl_natbib}
\appendix
\section{Additional Data and Model Details}\label{sec:appendix_data_model_settings}
In this section, we provide details regarding statistical testing, benchmark statistics, and miscellaneous details regarding our QA-based experiments. 
\subsection{Statistical Testing}
To determine whether the improvements on the SummaC benchmark are statistically significant, we perform significance tests using bootstrap resampling \cite{Efron}, following \citet{laban2021summac}.
We compare our best model to the best-performing model from prior work on a given subset of the benchmark.
We compare confidence intervals at significance levels of 0.05 and 0.01 and apply the Bonferroni correction \cite{bonferroni1935calcolo}.
Statistically significant differences at the 0.01 level exist between  \qafactevalnli~(supervised) and the best prior work on the FRANK subset and for the overall benchmark result. 
We do not see statistically significant differences on the other datasets in the benchmark. 
However, the statistically significant difference at the overall benchmark is notable; while other metrics may perform comparably or better on a given dataset, our metric demonstrates consistent good performance across datasets.
\subsection{Benchmark Statistics}
For completeness, we provide additional statistics for the SummaC benchmark in Table \ref{tab:summac_stats}.
Due to the exclusion of Omission and Addition as factual consistency errors in the Polytope dataset, our dataset contains benchmark replication contains many more positive examples for that dataset. 
For XSF, we restrict the dataset to those examples with labels for factual consistency with respect to the source, as opposed to more general factuality labels which take into account world knowledge, which results in fewer examples than the original SummaC benchmark. 
This is the same subset as was used in \citet{koto2021ffci}.
\par
Please see the following links for the licenses of the datasets and annotations: CGS\footnote{\url{https://tudatalib.ulb.tu-darmstadt.de/handle/tudatalib/2002}}, XSF\footnote{\url{https://github.com/google-research-datasets/xsum_hallucination_annotations\#license}}, FactCC\footnote{\url{https://github.com/salesforce/factCC/blob/master/LICENSE.txt}}, SummEval\footnote{\url{https://github.com/Yale-LILY/SummEval/blob/master/LICENSE}}.
We did not find licenses for the remaining datasets analyzed in our study. 
The intended uses of these licenses align with our use for research purposes.

\begin{table}[t!]
\resizebox{\columnwidth}{!}{\begin{tabular}{|c|c|c|c|}
\hline
    \textbf{Dataset}   & \textbf{\# Valid}  & \textbf{\# Test} & \textbf{\% Positive}    \\ \hline
    CGS         &    1281     &  400 & 49.7  \\  \hline
    XSF         &    996     &  996 & 9.4  \\  \hline
    Polytope    &    634     &  634 & 87.2  \\  \hline
    FactCC      & 931     &  503 & 85.8  \\  \hline
    SummEval    & 850     &  850 & 90.6  \\  \hline
    FRANK         & 671     &  1575 & 33.2  \\  \hline
                                  
\end{tabular}}
\caption{Statistics of the six datasets in the SummaC benchmark. We provide the number of validation and test set examples and the percentage of positive examples in the validation set.}
\label{tab:summac_stats}
\end{table}
\subsection{Model Parameters}
Ablation experiments started from a combination that provided good initial validation results and then swapped components. 
Running every combination of QA-based metric components is expensive. 
We experimented with running an ablation of the QA models with a 2nd-best performing answer selection component \textit{ALL}. 
This reduced all scores compared to using the NP Chunks component. 
This experiment supports our setup of keeping the best component constant when running ablations in order to determine the highest-performing combination of components, rather than experimenting with every combination.
\par
Inference for the MADE QA model is run using the average of the six MADE adapters' parameters. 
\par
For Question Filtering with the IsAnsweredSumm Filter, in addition to if the Electra-large QA model labels the question as unanswerable, if the \textit{F1} overlap score between the selected answer and the QA model output is less than 0.60, we remove this question. 
This filter was added only to IsAnsweredSumm and not IsAnsweredInput as answering questions based on the summary, from which the question was generated, should be an easy task.
We reached this threshold based on a qualitative analysis of model outputs, although this number could have also been further tuned on the validation set.  
\section{Additional Correlation Results}\label{sec:appendix_corr}
We provide additional correlation coefficients as a point of reference for future work. 
Instance-level correlations calculate the correlation between all instances, while the summary-level correlation computes the correlation between scores for each summary of the same input and then averages over inputs.  
Summary-level correlations are excluded for QAGS as this dataset does not contain annotations for multiple models, which is necessary to compute this score. 
\begin{table*}[t!]
\resizebox{\textwidth}{!}{\begin{tabular}{|c|c|c|c|c|c|c|c|}
\hline
    \textbf{Model Type}   & \textbf{Model Name} & \textbf{XSF} & \textbf{SummEval} & \textbf{FRANK-CNNDM} &  \textbf{FRANK-XSum} & \textbf{QAGs-CNNDM} &  \textbf{QAGs-XSum}    \\ \hline
    \multirow{2}{*}{Misc}         & BARTScore     & 0.25 & 0.34 & \textbf{0.54} & 0.14 & \textbf{0.68} & 0.17  \\
                                  & BLANC     & 0.07 & 0.20 & 0.33 & 0.06 & 0.30 & 0.03  \\ \hline
                                  & FactCC     &  0.05 & 0.37 & 0.41 & 0.05 & 0.49 & 0.26  \\ 
    \multirow{6}{*}{Entailment}   & BertScore-FFCI     & \textbf{0.45} & 0.26 & 0.34 & 0.15 & 0.50 & 0.20    \\ 
                                  & DAE     &  0.00 & 0.40 & 0.49 & 0.20 & 0.58 & -0.14    \\ 
                                  & ANLI     &  0.18 & 0.35 & 0.46 & 0.08 & 0.60 & 0.36  \\ 
                                  & MNLI     &  0.16 & 0.39 & 0.49 & 0.11 & 0.61 & 0.35  \\ 
                                  & DocNLI     &  0.01 & 0.34 & 0.11 & 0.21 & 0.21 & -0.38 \\ 
                                  & SCZeroShot     &  0.06 & 0.39 & 0.48 & 0.23 & 0.52 & 0.44 \\ \hline
    \multirow{2}{*}{QA}           & QuestEval  &  \textbf{0.43} & 0.33 & 0.47 & \textbf{0.24} & 0.45 & 0.24   \\ 
                                  & \qafacteval~    & 0.30 & \textbf{0.43} & \textbf{0.54} & \textbf{0.26} & \textbf{0.64} & \textbf{0.44}    \\ \hline
    \multirow{2}{*}{Learned}     & SCConv (synthetic)  &  0.19 & 0.41 & \textbf{0.54} & 0.22 & 0.04 & 0.04     \\ 
                                  & \qafactevalnli (synthetic)    &  0.16 & \textbf{0.47} & \textbf{0.60} & 0.21 & \textbf{0.64} & \textbf{0.47}     \\ \hline
\end{tabular}}
\caption{Instance-level Spearman correlation coefficients across factual consistency evaluation datasets. Metrics are divided into entailment-based, QA-based, and learned metrics that are fine-tuned on synthetic or supervised data. The two highest-correlated metrics for each dataset are shown in bold.}
\label{tab:corr_glob_spearman}
\end{table*}
\begin{table*}[t!]
\resizebox{\textwidth}{!}{\begin{tabular}{|c|c|c|c|c|c|c|c|}
\hline
    \textbf{Model Type}   & \textbf{Model Name} & \textbf{XSF} & \textbf{SummEval} & \textbf{FRANK-CNNDM} &  \textbf{FRANK-XSum} & \textbf{QAGs-CNNDM} &  \textbf{QAGs-XSum}    \\ \hline
    \multirow{2}{*}{Misc}         & BARTScore     & 0.17 & 0.27 & 0.42 & 0.12 & \textbf{0.55} & 0.14  \\
                                  & BLANC     & 0.05 & 0.15 & 0.25 & 0.05 & 0.24 & 0.02  \\ \hline
                                  & FactCC     & 0.03 & 0.29 & 0.31 & 0.04 & 0.38 & 0.21   \\ 
    \multirow{6}{*}{Entailment}   & BertScore-FFCI     & \textbf{0.31} & 0.20 & 0.25 & 0.12 & 0.39 & 0.16    \\ 
                                  & DAE     &  0.00 & 0.32 & 0.38 & 0.16 & 0.47 & -0.11   \\ 
                                  & ANLI     &  0.12 & 0.28 & 0.36 & 0.07 & 0.48 & 0.30  \\ 
                                  & MNLI     &  0.11 & 0.31 & 0.38 & 0.09 & 0.49 & 0.28   \\ 
                                  & DocNLI     &  0.01 & 0.27 & 0.08 & 0.17 & 0.17 & -0.31   \\ 
                                  & SCZeroShot     &  0.04 & 0.31 & 0.37 & 0.18 & 0.41 & 0.36  \\ \hline
    \multirow{2}{*}{QA}           & QuestEval  &  \textbf{0.30} & 0.26 & 0.36 & \textbf{0.20} & 0.35 & 0.20   \\ 
                                  & \qafacteval~    & 0.22 & \textbf{0.34} & \textbf{0.43} & \textbf{0.23} & \textbf{0.51} & \textbf{0.36}    \\ \hline
    \multirow{2}{*}{Learned}     & SCConv (synthetic)  &  0.13 & 0.33 & 0.42 & 0.18 & 0.03 & 0.03    \\ 
                                  & \qafactevalnli (synthetic)    &  0.11 & \textbf{0.37} & \textbf{0.47} & 0.17 & \textbf{0.51} & \textbf{0.38}     \\ \hline
\end{tabular}}
\caption{Instance-level Kendall correlation coefficients across factual consistency evaluation datasets. Metrics are divided into entailment-based, QA-based, and learned metrics that are fine-tuned on synthetic or supervised data. The two highest-correlated metrics for each dataset are shown in bold.}
\label{tab:corr_glob_kendall}
\end{table*}
\begin{table*}[t!]
\resizebox{\textwidth}{!}{\begin{tabular}{|c|c|c|c|c|c|c|c|}
\hline
    \textbf{Model Type}   & \textbf{Model Name} & \textbf{XSF} & \textbf{SummEval} & \textbf{FRANK-CNNDM} &  \textbf{FRANK-XSum}   \\ \hline
    \multirow{2}{*}{Misc}         & BARTScore     & 0.18 & 0.40 & 0.65 & 0.29   \\
                                  & BLANC     & 0.12 & 0.27 & 0.47 & 0.01   \\ \hline
                                  & FactCC     &  -0.02 & 0.39 & 0.40 & -0.07    \\ 
    \multirow{6}{*}{Entailment}   & BertScore-FFCI     & 0.21 & 0.37 & 0.44 & 0.19   \\ 
                                  & DAE     &  0.01 & 0.51 & 0.54 & 0.32    \\ 
                                  & ANLI     &  0.09 & 0.49 & 0.53 & 0.18   \\ 
                                  & MNLI     &  0.10 & 0.48 & 0.51 & 0.17     \\ 
                                  & DocNLI     &  0.00 & 0.52 & 0.21 & 0.47    \\ 
                                  & SCZeroShot     &  0.11 & 0.57 & 0.60 & \textbf{0.52}  \\ \hline
    \multirow{2}{*}{QA}           & QuestEval  &  \textbf{0.30} & 0.45 & 0.54 & 0.44    \\ 
                                  & \qafacteval~    & \textbf{0.24} & \textbf{0.64} & \textbf{0.68} & \textbf{0.53}    \\ \hline
    \multirow{2}{*}{Learned}     & SCConv (synthetic)  &  0.17 & 0.54 & 0.60 & 0.46    \\ 
                                  & \qafactevalnli (synthetic)    &  0.16 & \textbf{0.64} & \textbf{0.70} & 0.48    \\ \hline
\end{tabular}}
\caption{Summary-level Pearson correlation coefficients across factual consistency evaluation datasets. Metrics are divided into entailment-based, QA-based, and learned metrics that are fine-tuned on synthetic or supervised data. The two highest-correlated metrics for each dataset are shown in bold.}
\label{tab:corr_summ_pearson}
\end{table*}
\begin{table*}[t!]
\resizebox{\textwidth}{!}{\begin{tabular}{|c|c|c|c|c|c|c|c|}
\hline
    \textbf{Model Type}   & \textbf{Model Name} & \textbf{XSF} & \textbf{SummEval} & \textbf{FRANK-CNNDM} &  \textbf{FRANK-XSum}     \\ \hline
    \multirow{2}{*}{Misc}         & BARTScore     & 0.18 & 0.38 & \textbf{0.59} & 0.28   \\
                                  & BLANC     & 0.12 & 0.25 & 0.43 & 0.06   \\ \hline
                                  & FactCC     & 0.00 & 0.37 & 0.42 & -0.01    \\ 
    \multirow{6}{*}{Entailment}   & BertScore-FFCI     & 0.21 & 0.34 & 0.40 & 0.20   \\ 
                                  & DAE     &  0.00 & 0.40 & 0.47 & 0.30   \\ 
                                  & ANLI     &  0.10 & 0.39 & 0.47 & 0.17  \\ 
                                  & MNLI     &  0.08 & 0.38 & 0.48 & 0.15    \\ 
                                  & DocNLI     & -0.02 & 0.39 & 0.19 & 0.41   \\ 
                                  & SCZeroShot     & 0.11 & 0.41 & 0.51 & \textbf{0.50}    \\ \hline
    \multirow{2}{*}{QA}           & QuestEval  &  \textbf{0.27} & 0.35 & 0.47 & 0.45    \\ 
                                  & \qafacteval~    &  \textbf{0.22} & \textbf{0.45} & \textbf{0.59} & 0.47     \\ \hline
    \multirow{2}{*}{Learned}     & SCConv (synthetic)  &  0.16 & 0.43 & 0.55 & 0.44      \\ 
                                  & \qafactevalnli (synthetic)    & 0.17 & \textbf{0.47} & \textbf{0.63} & \textbf{0.49}   \\ \hline
\end{tabular}}
\caption{Summary-level Spearman correlation coefficients across factual consistency evaluation datasets. Metrics are divided into entailment-based, QA-based, and learned metrics that are fine-tuned on synthetic or supervised data. The two highest-correlated metrics for each dataset are shown in bold.}
\label{tab:corr_summ_spearman}
\end{table*}
\begin{table*}[t!]
\resizebox{\textwidth}{!}{\begin{tabular}{|c|c|c|c|c|c|c|c|}
\hline
    \textbf{Model Type}   & \textbf{Model Name} & \textbf{XSF} & \textbf{SummEval} & \textbf{FRANK-CNNDM} &  \textbf{FRANK-XSum}   \\ \hline
    \multirow{2}{*}{Misc}         & BARTScore     & 0.15 & 0.32 & \textbf{0.51} & 0.25  \\
                                  & BLANC     & 0.11 & 0.21 & 0.38 & 0.05  \\ \hline
                                  & FactCC     & 0.00 & 0.30 & 0.35 & -0.01     \\ 
    \multirow{6}{*}{Entailment}   & BertScore-FFCI     & 0.17 & 0.28 & 0.34 & 0.18    \\ 
                                  & DAE     &  0.00 & 0.33 & 0.41 & 0.27    \\ 
                                  & ANLI     &  0.08 & 0.32 & 0.41 & 0.16   \\ 
                                  & MNLI     &  0.07 & 0.31 & 0.41 & 0.14     \\ 
                                  & DocNLI     &  -0.01 & 0.32 & 0.17 & 0.37   \\ 
                                  & SCZeroShot     &  0.10 & 0.34 & 0.44 & \textbf{0.45}  \\ \hline
    \multirow{2}{*}{QA}           & QuestEval  &  \textbf{0.23} & 0.29 & 0.41 & 0.41    \\ 
                                  & \qafacteval~    & \textbf{0.19} & \textbf{0.37} & 0.51 & \textbf{0.45}   \\ \hline
    \multirow{2}{*}{Learned}     & SCConv (synthetic)  &  0.14 & 0.36 & 0.49 & 0.41    \\ 
                                  & \qafactevalnli (synthetic)    &  0.14 & \textbf{0.39} & \textbf{0.55} & 0.44 \\ \hline
\end{tabular}}
\caption{Summary-level Kendall correlation coefficients across factual consistency evaluation datasets. Metrics are divided into entailment-based, QA-based, and learned metrics that are fine-tuned on synthetic or supervised data. The two highest-correlated metrics for each dataset are shown in bold.}
\label{tab:corr_summ_kendall}
\end{table*}

\end{document}